\documentclass[conference]{IEEEtran}
\IEEEoverridecommandlockouts
\usepackage{cite}
\usepackage{amsmath,amssymb,amsfonts}
\usepackage{algorithmic}
\usepackage{graphicx}
\usepackage{textcomp}
\usepackage[dvipsnames]{xcolor}
\usepackage{url}
\usepackage{algorithm2e}
\def\BibTeX{{\rm B\kern-.05em{\sc i\kern-.025em b}\kern-.08em
    T\kern-.1667em\lower.7ex\hbox{E}\kern-.125emX}}

\begin{document}

\title{Title}


\title{Large Language and Text-to-3D Models for Engineering Design Optimization}

\author{\IEEEauthorblockN{Thiago Rios}
\IEEEauthorblockA{Honda Research Institute Europe \\
Offenbach am Main, Germany \\
thiago.rios@honda-ri.de}
\and
\IEEEauthorblockN{Stefan Menzel}
\IEEEauthorblockA{Honda Research Institute Europe\\
Offenbach am Main, Germany \\
stefan.menzel@honda-ri.de}
\and
\IEEEauthorblockN{Bernhard Sendhoff}
\IEEEauthorblockA{Honda Research Institute Europe \\
Offenbach am Main, Germany \\
bernhard.sendhoff@honda-ri.de}
}

\maketitle

\begin{abstract}
The current advances in generative AI for learning large neural network models with the capability to produce essays, images, music and even 3D assets from text prompts create opportunities for a manifold of disciplines. In the present paper, we study the potential of deep text-to-3D and text-to-text models in the engineering domain, with focus on the chances and challenges when integrating and interacting with 3D assets in computational simulation-based design optimization. In contrast to traditional design optimization of 3D geometries that often searches for the optimum designs using numerical representations, such as B-Spline surface or deformation parameters in vehicle aerodynamic optimization, natural language challenges the optimization framework by requiring a different interpretation of variation operators while at the same time may ease and motivate the human user interaction. Here, we propose and realize a fully automated evolutionary design optimization framework using Shap-E, a recently published text-to-3D asset network by OpenAI, in the context of aerodynamic vehicle optimization. For representing text prompts in the evolutionary optimization, we evaluate (a) a bag-of-words approach based on prompt templates and Wordnet samples, and (b) a tokenisation approach based on prompt templates and the byte pair encoding method from GPT4. Our main findings from the optimizations indicate that, first, it is important to ensure that the designs generated from prompts are within the object class of application, i.e. diverse and novel designs need to be realistic, and, second, that more research is required to develop methods where the strength of text prompt variations and the resulting variations of the 3D designs share causal relations to some degree to improve the optimization.
\end{abstract}



\section{Introduction}
The recent advances in building foundation models \cite{bommasani2022opportunities}, large language models (LLM) \cite{radford2019language}, and text-to-image models \cite{Rombach2021} have a major impact on a variety of fields, such as natural language processing and understanding, text and image generation, and human machine interaction. The maturity and ease of use of these novel models even lead to the adaptation of business models in some domains e.g., text writing, software development, and product design. The application of foundation models to engineering has been less discussed compared to other domains. Nevertheless, we see great potential in how large language models, text-to-image and text-to-3D models could be used in industrial engineering. Natural language interfaces between engineers and complex software systems in computational aided engineering could improve their usage and make them more accessible for younger engineers or for non-experts in general. Furthermore, \textit{text-to-X} approaches could improve the interaction between engineers and computer-aided design (CAD) and engineering (CAE) systems by offering new ways for generating designs and realizing design changes (images and 3D objects).

In computational engineering optimization text-to-3D generative models could be used as unique design representations. Building on former design data they would allow the exploration of the design space through language. From a general perspective, there are many alternative ways to describe a 3D object. All object representations rely on a number of parameters that are manipulated so that a certain object is realized. Traditional object representations are
spline curves and surfaces, which are e.g. parameterized by control points and knot points. Free-form deformations, which represent object changes, are particularly suitable for representing modifications of complex objects especially when combined with finite element/volume simulations \cite{PN3207}. CAD systems use complex and specific ways to represent objects, and computational engineering simulations typically rely on high resolution
meshes (e.g. triangulated mesh surfaces). These meshes are used for approximating physical equations like the Navier-Stokes equation for fluid dynamics calculations. The choice of the representation of an object depends on the how the representation is used in the computational engineering process. Recently, also deep learning based representations, such as (variational) autoencoders ((V)AEs) have been applied to generate point cloud designs \cite{Bronstein2017,Achlioptas2018}. By modifying the parameters of the latent space design variations are realized and processed in downstream applications like simulation-based optimization. Particularly, for engineering optimization, the representation largely determines the efficiency of the optimization and the quality of solutions that can be reached. 

\begin{figure}[htbp]
\centerline{\includegraphics{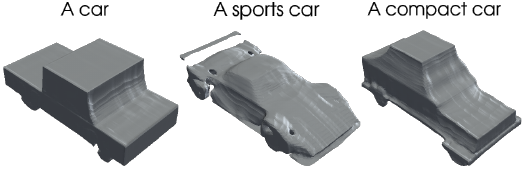}}
\caption{Examples of car shapes generated using Shap-E based on the prompts \textit{``A car"}, \textit{``A sports car"}, and \textit{``A compact car"}.}
\label{fig:car_samples}
\end{figure}

Using a representation for design optimization that describes a 3D-object with natural language offers a very different and unique approach to generate 3D shapes. Even though the benefit of having a very intuitive description of objects and their changes is evident, the quality of text-to-3D models in the context of design optimization is unknown and needs to be assessed. Here, not only the meaningfulness of designs for a given text prompt is of interest, but also the relation between prompt variations and resulting design variations. Since the models rely on curated data sets, it is important to understand in how far "novel" designs can be generated. By "novel", here we refer to designs which are different from the training data, yet realistic for a given application. In an engineering design optimization framework, we need to ensure that the generated designs are consistent with the optimization target, e.g. a prompt including the word "car" should result in a vehicle-like geometry, while (subtle) variations like additional attributes or adjectives result in recognizable variations, e.g., a "compact car" should be different from a "sports car" \mbox{(Fig. \ref{fig:car_samples})}. 

It is also important to get insights into the transition between generated designs of similar shapes to "hallucinated" designs to identify if a generated design is "novel", maybe part of a larger design group, or if it is an outlier or artefact. Therefore, in the present paper, we discuss the potential of multi-modal large language models as representations of 3D objects for simulation-based engineering design optimization from a practitioners perspective.

Firstly, we propose a fully automated computational 3D design optimization framework for vehicle development that integrates the recently published Shap-E \cite{Jun2023} as a \mbox{text-to-3D} generative model. Secondly, we analyze our optimization results to identify benefits for interacting with engineering tools through natural language models. For example, replacing standard representations, such as spline curves/surfaces or free form deformations \cite{PN3207, Rios2021c}, with text prompts has direct implications not only on the quality of the automatically generated models but also potentially improves the interpretation of the characteristics of the optimized design. Although we use automotive applications in this paper, the methods and conclusions equally apply to other design optimization problems such as from aviation, or marine.

The remainder of this paper is outlined as follows: In Section \ref{rev_lit}, we discuss deep learning models for evolutionary design optimization, like generative models for text-to-3D tasks and prompt engineering. Section \ref{methods} details our proposed design optimization framework with a focus on the different approaches for representing text prompts. In Section \ref{experiments}, we demonstrate the application of our framework to a simulation-based design optimization for the minimization of vehicle drag coefficients and discuss the results of our experiments. \mbox{Section \ref{conclusion}} concludes the paper.





\section{Literature review}\label{rev_lit}
In this section, we will describe evolutionary design optimization in engineering with learning-based shape representations. Then, we highlight the current state in text-to-3D and text-to-image generative models followed by approaches for prompt engineering. 

\subsection{Geometric deep learning for 3D vehicle optimization}
The optimization of the shape of 3D objects is an important step in product design. For automotive engineering, the shape of cars is optimized e.g for fuel efficiency, or crash safety. A computational engineering optimization framework typically consists of the shape parameterization, i.e., the representation(s), the optimization algorithm for modifying the shape parameters and simulation tools for determining the design performance. Many different approaches for representing shapes have been proposed and all have certain advantages and disadvantages. It is important to note that for almost all realistic engineering frameworks the shape is represented at least two, realistically even three times: the first representation is used for the optimization algorithm, i.e., the optimization modifies the parameters of this representation. The second representation is used for the simulation algorithm, which usually uses a high resolution mesh for solving differential equations. The third representation is finally used for the actual manufacturing process. When we describe shape representations in this paper, we mostly refer to the first one, the one whose parameters are subject to optimization. However, the transition between the different representations is of high practical relevance and should not be overlooked in the computational engineering optimization framework.

The introduction of geometric deep learning architectures \cite{Bronstein2017} enabled the development of 3D deep-generative models for engineering tasks. Most of the currently available works focus on learning compact representations of 3D objects for shape generation and performance prediction. In \cite{Umetani2017}, Umetani proposes a system for generating 3D car designs and for predicting the corresponding aerodynamic performance. The system is based on a deep autoencoder architecture, where, by manipulating the values of the learned latent representation, the user quickly generates 3D car designs and obtains an estimate of the aerodynamic drag of the shape. 

Rios \textit{et. al} build upon a 3D point cloud autoencoder \cite{Achlioptas2018} and propose an automated framework for car design optimization based on evolutionary algorithms \cite{Rios2021a} and multi-task optimization methods \cite{Rios2021b}. The authors show that the point-based networks learn variations of local geometric features better than global transformation methods, \textit{e.g.}, principal component analysis, which yields better performance in nonlinear design optimization problems. Saha \textit{et. al} evaluate point-based (variational) autoencoders ((V)AEs) with respect to their shape-generative and performance prediction capabilities \cite{Saha2020, Saha2021}. One target is to understand whether VAEs are able to generate novel yet realistic designs. The authors claim that, compared to the standard architecture of variational autoencoders, a proposed regularization of the latent space enables smoother design variations with less artifacts, which is beneficial for 3D shape synthesis.
However, this comes at the expense of a reduced accuracy of surrogate models, which learn to predict the performance of 3D designs using the latent space representation as input. 

Learned design representations have a number of advantages compared to traditional representations like splines or free form deformation. For complex shapes, the parameterization of traditional representations requires an experienced engineer who balances the freedom of design variations with the dimensionality of the search space. It is not uncommon that during the search process these representations have to be adapted in order to increase the design flexibility of a certain part of the 3D object. Apart from the additional effort and the dependence on the expertise of the engineer, the freedom of generating truly novel and unique designs is restricted by the choice of the representation, i.e., by the inherent assumption of the engineer where in the 3D object successful variations are most likely to be realized. Of course learned design representations also have their challenges. Firstly, the flexibility of the representation depends on the variations in the data set that is used for training. The capability to extrapolate from the 
seen data is unclear and unpredictable. Secondly, representations learned by deep neural networks are difficult to interpret. Therefore, it is impossible to interfere during the optimization process or to insert some design preferences as it is the target of cooperative engineering design systems. For both, traditional and learned representations another challenge is the automated generation of polygonal meshes to perform simulations.


\subsection{Deep generative models for text-to-3D assets}
Recently, generative models, namely Point-E \cite{Nichol2022} and \mbox{Shap-E} \cite{Jun2023}, have been proposed by OpenAI which offer first capabilities to generate 3D objects from text prompts. As stated above, the potential to generate 3D designs from natural language enables alternative ways for industrial design optimization and interactive design optimization. Point-E is a generative model that produces 3D point clouds from text prompts. It first generates a single synthetic view using a \mbox{text-to-image} diffusion model, here GLIDE \cite{nichol2022glide}, which is followed by a second diffusion model which is conditioned on the generated image to calculate the 3D point cloud. Without going further into detail, text-to-image models, such as GLIDE, Dall-E \cite{ramesh2021zeroshot}, Midjourney and Stable Diffusion \cite{Rombach2021}, rely on diffusion models that are trained on large data sets of annotated images and can generate high-quality images from noisy images. In the second step of Point-E, Nichols et al. proposed a novel transformer architecture to include RGB colors of each point and trained the network on (image, 3D) pairs. In a downstream process, they used an upscaler on the point clouds and traditional (limited) meshing methods for rendering purposes. 

Shap-E \cite{Jun2023} builds upon Point-E and generates 3D meshed objects, which is often required for rendering simulation-based applications in engineering. The network is trained on a similar data set as Point-E, which comprises refined point clouds (16K points) and rendered views of 3D objects from multiple classes that are paired to text prompts. Similar to Point-E, the network relies on diffusion operations in the latent space representations, but learns to generate implicit representations (signed distance functions) of 3D objects, which are utilized in a differentiable implementation of the marching cubes 33 algorithm to generate polygonal meshes.

Besides Shap-E, there are several other approaches that could be integrated into our proposed design optimization framework. However, these models are currently not accessible. Dreamfusion \cite{poole2022dreamfusion} utilizes a pretrained 2D text-to-image diffusion model to perform text-to-3D synthesis to avoid the need for large scale 3D labelled data. By optimizing a randomly-initialized 3D Neural Radiance Field model (NeRF), which consists of a volumetric raytracer and a multilayer perceptron, and using their proposed Score Distillation Sampling loss function, the authors generated coherent 3D scenes from text prompts. With the development of Magic3D \cite{lin2023magic3d} the authors want to overcome some of the drawbacks of Dreamfusion like (a) extremely slow optimization due to the NeRFs and (b) low-resolution images. They optimize neural field representations on a coarse level (color, density, and normal fields) and extract a textured 3D mesh from the density and color fields. In a final step, they use a high-resolution latent diffusion model to generate high-quality 3D meshes with detailed textures. Get3D \cite{gao2022get3d} utilizes 3D generative models that synthesize textured meshes for direct usage in 3D rendering engines. They combine a differentiable explicit surface extraction method based on signed distance fields to get a 3D mesh topology and a differentiable rendering technique to learn the texture of the surface. With Gaudi \cite{bautista2022gaudi}, the authors propose a generative model capable of capturing the distribution of complex and realistic 3D scenes that can be rendered from a moving camera. 

In addition, it should be noted that recently first attempts have also been made to utilize chatGPT [2] for writing python scripts for blender\footnote{https://www.blender.org} to generate 3D scenes using text prompts.

\subsection{Prompt engineering and optimization}
Naturally, as text prompts become more and more popular as natural language interfaces for \textit{text-to-X} models, prompt engineering \cite{reynolds2021} and prompt optimization gets high attention. Besides choosing an adequate model for the required theme, prompt engineering in text-to-image applications allows to increase the quality of the generated images. Referring in the prompt to famous artists like Picasso, van Gogh, Warhol, etc., or to art styles like photography, oil painting, sketch etc., influences the result significantly. Prompt engineering can also be used as a negating description, e.g., bad anatomy, wrong hands, etc.
In \cite{Martins2023}, the authors utilize an interactive evolutionary approach to find optimal prompts in a text-to-image scenario. They create meta prompts to represent spaces of prompts and then use a genetic algorithm to improve the prompt based on feedback interactively provided by the user on resulting image qualities. In \cite{wong2023}, the authors apply a multi-objective evolutionary optimization to evolve user prompts for image generation for improved consideration of user preferences. The authors claim that especially the usage of the conditional generative model as a kind of mutation guidance is novel.

In \cite{arechiga2023dragguided}, the authors build a surrogate model, which has been trained on pairs of 2D vehicle renderings and associated drag coefficients resulting from CFD simulation, and integrate it into a text-to-image model, here Stable Diffusion. As a result, their drag-guided diffusion model is capable to generate vehicle images that are aerodynamically efficient.  







\section{Methods}\label{methods}
\begin{figure}[htbp]
\centerline{\includegraphics{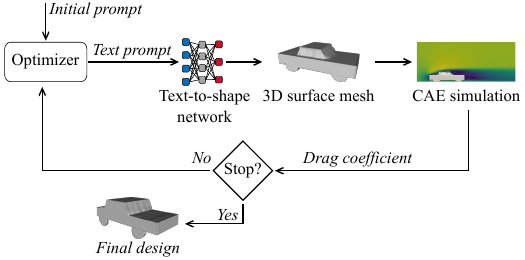}}
\caption{Workflow of the method utilized in our experiments for optimizing 3D designs using a text-to-shape generative DNN and CAE simulations.}
\label{fig:opt_flow_chart}
\end{figure}

For the evaluation of a text-to-shape deep neural network as a 3D shape-generative model in the context of a simulation-based design optimization, we propose to adapt an evolutionary design optimization framework, which is frequently applied in engineering optimization applications using traditional representations \cite{JinTEC,PN3207}, when gradients of the performance function are not easy or impossible to calculate. We adapt the framework by integrating the text-to-3D generative model for mapping the optimization parameters to 3D shapes, see \mbox{Fig. \ref{fig:opt_flow_chart}}.


The framework consists of three main components: an optimizer for directed parameter variation, a text-to-shape model for design generation, and a simulation tool for design performance computation. Based on an initial text prompt, the evolutionary optimizer generates a population of prompts by randomly modifying the initial text. Then, the prompts are read by the shape generation model, which generates corresponding 3D shapes represented as polygonal meshes, which are post-processed and used for the computational simulation. Here we use OpenFOAM for computational fluid dynamics calculations to determine aerodynamic efficiency. Finally, based on a set of convergence criteria such as convergence of step size adaption or maximum iteration counts, either the optimization loop stops and yields the best performing shape, or iterates by generating a new population of prompts based on applying evolutionary operators to the individuals with best performance.

\subsection{Design domain}
We consider two approaches for representing the text prompts: Bag-of-words and tokenisation. In the bag-of-words approach, we define a prompt template and sets of words which can be utilized to complete the template. Since we target the aerodynamic optimization of car designs, we defined the template as
$$ \text{A} <adjective> \text{car in the shape of} <noun>,$$
where the \textit{adjective} and \textit{noun} are sampled from the set of words available in Wordnet \cite{Fellbaum2019}. Furthermore, we encoded the words based on the similarity metrics proposed by Wu \& Palmer (WUP) \cite{Wu1994} with respect to the reference words ``fast" and ``wing", for the \textit{adjective} and \textit{noun}, respectively. For both words we would expect a resulting car with low drag coefficients. Hence, the optimization algorithm yields individuals that encode the distance to the reference words and the prompt is reconstructed by recovering the words in the sets with the most similar distance values. Furthermore, since the sets comprise only words that exist in the English grammar, the generated prompts are human-readable, even though some of the adjective-noun combinations lead to a prompt with counter-intuitive semantic interpretation.

Similarly, in the tokenisation approach, we also generate designs by modifying a prompt template: 
$$\text{A car in the shape of} <string>. $$
However, instead of sampling \textit{string} from a predefined set of words, we utilize the same byte pair encoding method as in GPT-4 \cite{openai2023gpt4} to generate the text for completing the prompt template. As the tokens are represented by integers, we directly utilize the values of the tokens as design variables. The main difference with respect to the bag-of-words approach is that the tokenisation allows us to generate \textit{strings} with any combination of characters. Hence, although we expect this method to often generate illegible strings, the tokenisation approach allows us to verify the robustness of the network against changes in the prompt and the influence of particular parts of words on the generated designs.

\subsection{Evolutionary optimization}\label{sec:opt_settings}
Since gradient information is often challenging to compute from CAE simulation models and in order to cope with multi-modal quality functions, evolutionary algorithms have been frequently applied to complex engineering optimization problems. Apart from being gradient-free, the search mechanisms of evolutionary methods cope with design variables encoded as continuous or binary variables, and can avoid local minima. In our experiments, we utilize the covariance matrix adaptation evolutionary strategy (CMA-ES) \cite{6790628} to optimize the designs. CMA-ES is an algorithm that has been successfully applied to engineering optimization before and can reach good results even for small population sizes and a limited number of generations. Furthermore, it is relatively robust with regard to the standard settings of the hyperparameters \cite{Baeck1996}. 

We set the population size $\lambda$ to 10, the number of parents $\mu$ to 3, and the maximum number of iterations to 100. The population is initialized with the following prompt 
$$\text{A } fast \text{ car in the shape of a } wing$$
for the optimizations based on the bag-of-word and with
$$\text{A car in the shape of a } wing$$
for the token representations, respectively. Furthermore, as CMA-ES assumes continuous variables, we generate the token values by approximating the generated parameter values to corresponding nearest integers, and limit the values to available range of tokens ($[0,32768)$). In both cases, we use a $(\mu, \lambda)$ strategy .

\subsection{Text-to-3D generative model}
In our experiments, we utilize a pre-trained version of Shap-E \cite{Jun2023} as text-to-3D generative model\footnote{The network architecture and weights are available at https://github.com/openai/shap-e}. The network was trained on an extended version of the dataset utilized in \cite{Nichol2022}, which comprises millions of 3D objects from different classes represented as point clouds (16K points) and by renderings. Hence, the utilized version of Shap-E generates objects from different classes and is not specialized on car designs. Also, we utilize the same network hyperparameters as proposed in the literature \cite{Jun2023}  and vary only the batch size according to the objectives of the experiment.

Since the generative process of Shap-E is probabilistic, i.e., the network generates slightly different shapes from the same prompt, we fixed the seed for random number generation to the same values for all our experiments. Furthermore, we generate 300 designs by feeding the prompt ``A car" to Shap-E to compute baseline performance metrics, such as the aerodynamic drag coefficient, which we utilize to evaluate the performance of the optimization framework. Furthermore, as Shap-E generates 3D shapes with different orientations, we re-align the designs assuming that the largest overall dimension corresponds to the length of the car ($x$-axis) and the smallest dimension corresponds to the height ($z$-axis).

\subsection{Aerodynamics simulation model}
The simulation framework utilized in the experiments is a direct adaptation of the available tutorial on aerodynamics simulation of a motorcycle using OpenFOAM\footnote{https://www.openfoam.com/} \cite{Dubey2020}. The only modification is on the fluid domain, where the motorcycle shape is replaced by the car geometries generated during the optimization. Further meshing and simulation settings, such as boundary conditions, are kept the same as proposed in the tutorial.

Regarding the computational settings, we perform the simulations in parallel using 12 processors on a single machine with two CPUs Intel\textregistered Xeon\textregistered Silver, clocked at 2.10 GHz and 196MB of RAM. The same machine is also equipped with 2 GPUs NVidia\textregistered Quadro\textregistered RTX 8000 (48 GB each) that are utilized to perform the computations for shape generations using Shap-E.

\section{Experiments and discussion}\label{experiments}
In this section, we discuss the experiments performed to evaluate Shap-E as a shape-generative model for evolutionary engineering design optimization. We first discuss the computation of the baseline performance metrics, followed by the optimization cases for each type of representation.

\subsection{Baseline performance metrics}
In our experiments, we define performance of the designs as the aerodynamic drag coefficient ($c_d$). As the shape-generative model generates different 3D designs from a same input text prompt, we defined the baseline performance of a car design generated by the network based on a data set of 300 shapes (batch size=300) generated from the prompt ``\textit{A car}". Furthermore, to verify the consistency of the simulation model, we also computed the length, height, width and projected frontal area of the generated shapes and verified their relation to the obtained $c_d$ values.

\begin{figure}[htbp]
\centerline{\includegraphics{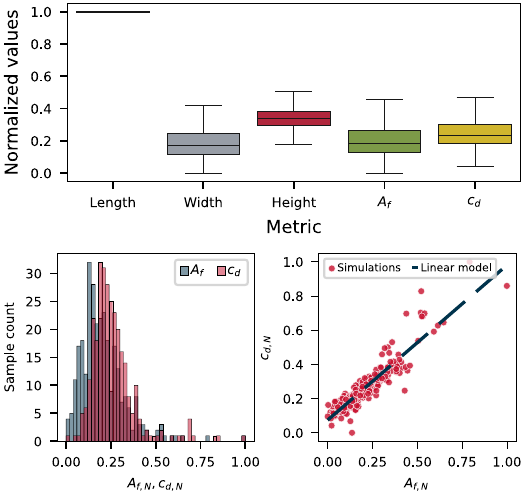}}
\caption{Visualization of the obtained distributions for the normalized performance measures (top). Correlation between the normalized projected frontal area $|A_f|$ and computed drag coefficient $c_{d,N}$ (bottom).}
\label{fig:baseline_distributions}
\end{figure}

By visualizing the obtained distributions of the selected metrics (Fig. \ref{fig:baseline_distributions}), we observe that the length of the generated designs is nearly identical for all designs. This effect can be explained by the normalization of the training data, which is a common practice in developing machine learning. In our studies, we assume that the projected frontal area has the largest impact on the drag coefficient of the designs. Thus, the optimization is unlikely to be affected by the similarity of the car lengths. Furthermore, we obtained similar distributions for the projected frontal area $A_f$ and the drag coefficient $c_d$, which is to be expected. By plotting $c_d$ as a function of $A_f$, we also observe that both are linearly correlated (R-squared of 0.8409), which indicates that the simulation settings are coherent with the physical phenomena. For this and the following analyses, the values of the performance measures are normalized based on the span of values of the baseline set (Eq. \ref{eq:norm}).

\begin{equation}\label{eq:norm}
    c_{d.N} =\frac{c_d}{max(c_{d,baseline})-min(c_{d,baseline})}
\end{equation}

\begin{figure}[htbp]
\centerline{\includegraphics{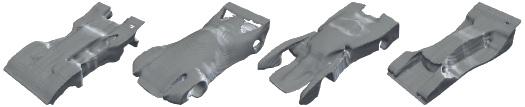}}
\caption{Examples of 3D designs obtained with the prompt "A \textit{fast} car in the shape of \textit{wing}" for computing the initial performance metrics.}
\label{fig:aero_design_0}
\end{figure}

Additionally, as a reference for the initial performance, we generated 50 designs by feeding the prompt 
$$\text{A } fast \text{ car in the shape of a }wing$$
to Shap-E (Fig. \ref{fig:aero_design_0}) and ran the simulations with the same CFD framework as applied to the baseline shapes. For this set of shapes, we obtained a mean 
normalized $c_d$ of $0.46 \pm 0.08$ for a confidence interval of 95\%.

\subsection{Prompt-based evolutionary design optimization}
\begin{figure}[htbp]
\centerline{\includegraphics{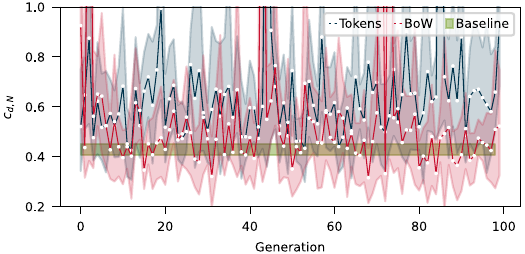}}
\caption{Baseline and mean of the population $c_d$ value during the optimization. The translucent areas represent a confidence interval of 95\% based on the population data for each generation.}
\label{fig:opt_converg}
\end{figure}

In this set of experiments, we utilize the CMA-ES to optimize 3D car shapes by modifying text-prompt templates. For the selected optimization settings (Section \ref{sec:opt_settings}), both, the bag-of-words (BoW) and tokenisation approaches yield slow convergence ratios and highly-oscillating population performances throughout the optimization (Fig. \ref{fig:opt_converg}). However, compared to the tokenisation approach, the BoW-based optimization generated designs with slightly lower $c_d$, which indicates that the higher degree of freedom of the tokenisation representation cannot be exploited by the optimization. 

\begin{figure}[htbp]
\centerline{\includegraphics{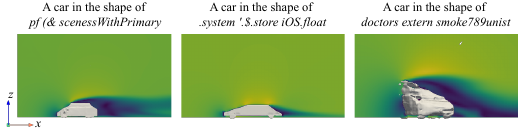}}
\caption{Examples of 3D shapes generated during the optimization based on the tokenisation approach and the corresponding velocity field obtained in the simulations. Brighter colors indicate higher velocity.}
\label{fig:opt_samples}
\end{figure}

One of the potential causes for the noisy behavior is the multi-modality of the quality landscape. For the tokenisation representation, even though the byte pair encoding method allows to generate a wide range of words from integers, the random variation of the tokens is prone to generate unintelligible expressions (Fig. \ref{fig:opt_samples}). Therefore, the generated shapes are predominately similar to car designs with counter-intuitive modifications, which results in performance values close to the baseline with some random variations. However, the tokens also generate some chunks of words with semantic interpretation, from which Shap-E creates designs with mixed features, e.g., a prompt containing the chunk ``smoke" potentially leads to a cloud-like car design (Fig. \ref{fig:opt_samples}, right).

\begin{figure}[htbp]
\centerline{\includegraphics{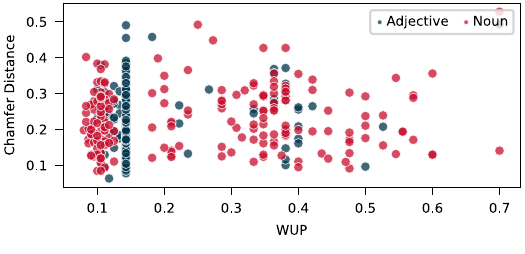}}
\caption{Visualization of the Chamfer distance as a function of the Wu \& Palmer metric for shapes generated with different combinations of adjectives and nouns with respect to designs generated by using ``\textit{a car}" as a text prompt.}
\label{fig:similarity}
\end{figure}

In the BoW approach, the optimization only generates prompts with intelligible words and, thus, provides a more intuitive relation between the prompts and design properties. We evaluate the prompt-to-shape relation by computing the WUP measure between 300 adjectives and nouns randomly sampled from Wordnet with respect to the word "car", and the Chamfer distance \cite{Fan2017} between the shapes generated by feeding only each of the sampled words to Shap-E and the design obtained with the prompt "car". By visualizing the obtained values (Fig. \ref{fig:similarity}), we observe that the samples are clustered around certain WUP values and spread over a wide range of Chamfer distance values. 

\begin{figure}[htbp]
\centerline{\includegraphics{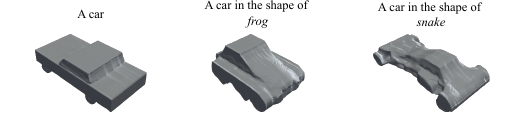}}
\caption{Visualization of the geometries generated by similar prompt modifiers. The WUP values of ``snake" and ``frog" with respect to ``car" are 0.35 and 0.36, respectively, but the generated designs are significantly different.}
\label{fig:similarity_geoms}
\end{figure}

This behavior is explained by the characteristics of the WUP metric, which is based on the depth of the compared terms in the Wordnet taxonomy. Hence, words that belong to the same semantic class but that describe geometrically distinct objects (e.g., ``snake" and ``frog" are animals, but physically very different) yield a similar WUP value and high variation in the geometric properties (Fig. \ref{fig:similarity_geoms}). Furthermore, this characteristic hinders the mapping of the selected design parameters to the performance measure since, by definition, mathematical functions map each sample in the domain to exclusively one element in the codomain.


\begin{figure}[htbp]
\centerline{\includegraphics{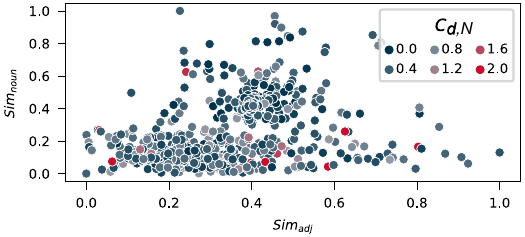}}
\caption{Visualization of the obtained $c_d$ for the samples generated during the optimization with the BoW approach as a function of the similarity of the adjectives and nouns with respect to the reference words (``\textit{fast}" and ``\textit{wing}").}
\label{fig:opt_landscape}
\end{figure}

Consequently, the optimization landscape becomes more complex and challenging for the optimization algorithm. 
In the BoW approach (Fig. \ref{fig:opt_landscape}), we observe that samples with distinct performance values overlap at different points in the design space, and that the landscape lacks a smooth trend for the $c_d$ value, which confirms our hypothesis. Furthermore, the complexity of the landscape also justifies the oscillating behavior of the populations' performance over the generations that was observed in the optimizations.

\begin{figure}[htbp]
\centerline{\includegraphics{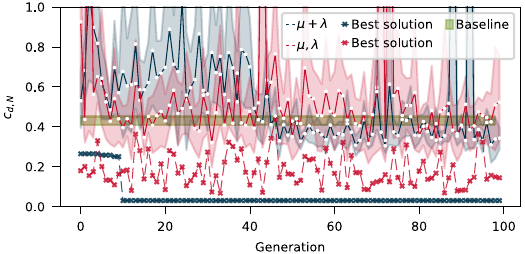}}
\caption{Mean and minimum normalized $c_{d,N}$ values of the population obtained during the optimizations using the $(\mu,\lambda)$ and $(\mu+\lambda)$ strategies. The translucent areas represent a confidence interval of 95\% based on the population data for each generation.}
\label{fig:opt_mu+lambda}
\end{figure}

In a second experiment, we changed the evolutionary strategy to $(\mu+\lambda)$ and carried out the optimization using the BoW approach since it seemed the more promising approach compared to the tokenisation representation. Since the best performing designs are always kept in the population, we expect the mean performance of the population to oscillate less and converge to lower $c_d$ values - of course at the expense of possibly getting tracked in local minima. Indeed the elitist $(\mu+\lambda)$ strategy is able to find designs with smaller $c_{d,N}$ values (Fig. \ref{fig:opt_mu+lambda}), but the variance throughout the optimization is comparable to the $(\mu,\lambda$) strategy. This can also be attributed to the lack of correlation between word similarity and geometric distance, as we already discussed.

\begin{figure}[htbp]
\centerline{\includegraphics{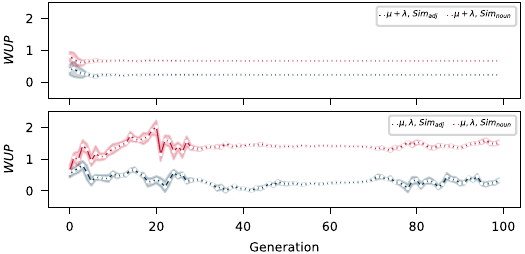}}
\caption{Mean word similarity metrics obtained over the generations for the noun and adjective in the $(\mu,\lambda)$ and $(\mu+\lambda)$ optimizations. The colored bands indicate a confidence interval of 95\% for each metric. }
\label{fig:opt_mu+lambda_params}
\end{figure}

Our interpretation of the results is also supported by the convergence of the design parameters over the generations (Fig. \ref{fig:opt_mu+lambda_params}). We observe that the variance of the WUP values decreases and stabilizes over the generations, particularly in the $(\mu+\lambda)$ scenario, which is to be expected for the elitist strategy.

\begin{figure}[htbp]
\centerline{\includegraphics{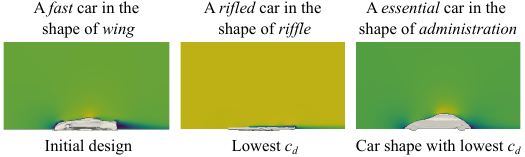}}
\caption{Initial design, shape with lowest $c_d$ and best performing car design obtained in the optimization with the $(\mu+\lambda)$ strategy. The colors indicate the magnitude of the fluid velocity, where brighter colors indicate higher velocity.}
\label{fig:opt_designs}
\end{figure}

For verification, we visually inspected the initial and best performing designs obtained in the $(\mu+\lambda)$ scenario (Fig. \ref{fig:opt_designs}). We observe that the design with lowest normalized $c_d$ value (0.02) is similar to a thin tube, which is potentially caused by the words "rifled" and "riffle" in the prompt. The latest and best performing car design has a normalized $c_{d,N}$ of 0.15, which is significantly better than the initial design.

\begin{figure}[htbp]
\centerline{\includegraphics{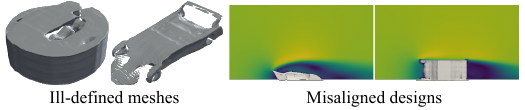}}
\caption{Examples of generated designs with mesh artifacts (left) and inconsistent orientation with respect to the CFD simulation framework (right).}
\label{fig:opt_limitations}
\end{figure}

To summarize, we show in our experiments that optimizing 3D designs based on text prompts using automated evolutionary optimization algorithms and CFD simulations is feasible. The proposed framework maps words to a numerical design space, which enables the CMA-ES algorithm to sample solutions, and finally generate a design with lower normalized drag coefficient than the initial shape (from $0.46$ to $0.15$). Yet, as the utilized numerical representation of text prompts is neither canonical nor a proper mathematical function, the proposed system is prone to generate designs with very distinct geometric properties from similar input values, which reduces the performance of the optimization. Furthermore, other inconsistencies in the geometric representation of the designs, such as mesh quality and orientation of the car shapes (Fig. \ref{fig:opt_limitations}) also increase the noise in the performance values and mislead the search for best performing designs. 

\section{Conclusion and Outlook}\label{conclusion}
In the present paper, we propose an evolutionary design optimization framework, which uses a text-to-3D generative model as a representation for mapping text prompts to 3D geometries. The generated shapes are simulated using a standard CFD program and evaluated for their drag value. We used two approaches for encoding the text prompts: (a) a bag-of-words approach utilizing Wordnet and (b) a tokenisation approach based on GPT4. The optimization algorithm adapts these prompts for finding 3D shapes with optimal aerodynamic performance, i.e. minimum drag coefficients. The primary targets of our studies are to explore the capability of the text-to-3D model embedded in an evolutionary engineering design optimization for (i) generating novel, yet realistic designs, and (ii) for representing a meaningful relation between prompt variations and design variations. 

Our results show that the text-to-3D models can be used as an alternative representation in engineering design optimization. Even though the optimization performance is substantially lower compared to traditional or autoencoder representations due to the yet not fully known text-to-3D model characteristics, it opens up many novel possibilities for the interaction with the engineer. Indeed, the generative aspect of the text-to-3D models, could address many open issues (like co-exploration of design spaces, qualitative description of design variations) in engineering design optimization and enable new processes in industrial designs. Nevertheless, in the set-up that we used in our experiments the generation of designs may also be error-prone resulting in non-car like geometries. There are several possible reasons like the unknown relation between the strength of a text prompt variation and the resulting design variation. Particularly, the adaptation of strategy parameters, which is important for the workings of the CMA-ES is likely to be severely affected by this complex relation between the measures in both spaces. Possibly other optimization methods like Different Evolution might be a better choice. At the same time,
the coarse and qualitative description of designs with the text-to-3D representation also has a potential to be more explorative and (possibly together with the engineer) might be able to identify truly novel regions in the design space. 
However, the large language models, which are an integral part of the text-to-3D representations rely on a significant element of randomness in the interpretation of the text prompt. Therefore, the design that corresponds to the textual description "car'' also has an inherent variability. The influence of this variability on the optimization process is not known. However, it can be interpreted as noise in the representation, which will very likely have an effect.   

This study is only a starting point in the research on using large language models and text-to-3D representations in engineering design optimization. There are many conceivable next steps to explore. One is the introduction of a classifier into the optimization, which identifies non-car like shapes and excludes them from optimization. Another is to augment the semantic distance measure between the different textual descriptions with a measure that relates more to the geometric difference. Last but not least, the exploitation of the interpretability of the representation in the interaction with the engineer is an exciting and novel field of research.



\bibliographystyle{./IEEEtran}
\bibliography{References}

\end{document}